\documentclass[
]{ceurart}

\usepackage{tcolorbox}

\begin{document}

\copyrightyear{2025}
\copyrightclause{Copyright for this paper by its authors.
  Use permitted under Creative Commons License Attribution 4.0
  International (CC BY 4.0).}

\conference{ROMCIR 2025: The 5th Workshop on Reducing Online Misinformation through Credible Information Retrieval (held as part of ECIR 2025: the 47th European Conference on Information Retrieval), April 10, 2025, Lucca, Italy}


\title{Towards Automated Fact-Checking of Real-World Claims: Exploring Task Formulation and Assessment with LLMs}

\author[1,2]{Premtim Sahitaj}[%
orcid=0000-0003-3908-5681,
email=sahitaj@tu-berlin.de,
]
\cormark[1]

\address[1]{Technische Universität Berlin, Quality and Usability Lab, Berlin, 10587, Germany}
\address[2]{Deutsches Forschungszentrum für Künstliche Intelligenz, Speech and Language Technology Lab, Berlin, 10559, Germany}

\author[3]{Iffat Maab}[%
email=maab@nii.ac.jp,
]

\address[3]{National Institute of Informatics, Digital Content and Media Sciences Research Division, Tokyo,  101-8430, Japan}

\address[4]{Harz University of Applied Sciences, Faculty of Automation and Computer Science, Wernigerode, 38855, Germany}

\author[3]{Junichi Yamagishi}[%
email=jyamagis@nii.ac.jp,
]

\author[4]{Jawan Kolanowski}[%
email=u37871@hs-harz.de,
]

\author[1,2]{Sebastian Möller}[%
orcid=0000-0003-3057-0760,
email=sebastian.moeller@tu-berlin.de,
]

\author[1,2]{Vera Schmitt}[%
orcid=0000-0002-9735-6956,
email=vera.schmitt@tu-berlin.de,
]
\cortext[1]{Corresponding author.}

\begin{abstract}
Fact-checking is necessary to address the increasing volume of misinformation. Traditional fact-checking relies on manual analysis to verify claims, but it is slow and resource-intensive. This study establishes baseline comparisons for Automated Fact-Checking (AFC) using Large Language Models (LLMs) across multiple labeling schemes (binary, three-class, five-class) and extends traditional claim verification by incorporating analysis, verdict classification, and explanation in a structured setup to provide comprehensive justifications for real-world claims. We evaluate Llama-3 models of varying sizes (3B, 8B, 70B) on 17,856 claims collected from PolitiFact (2007–2024) using evidence retrieved via restricted web searches. We utilize TIGERScore as a reference-free evaluation metric to score the justifications. Our results show that larger LLMs consistently outperform smaller LLMs in classification accuracy and justification quality without fine-tuning. We find that smaller LLMs in a one-shot scenario provide comparable task performance to fine-tuned Small Language Models (SLMs) with large context sizes, while larger LLMs consistently surpass them. Evidence integration improves performance across all models, with larger LLMs benefiting most. Distinguishing between nuanced labels remains challenging, emphasizing the need for further exploration of labeling schemes and alignment with evidences. Our findings demonstrate the potential of retrieval-augmented AFC with LLMs.
\end{abstract}


\maketitle
\section{Introduction}

Misinformation, whether spread inadvertently or with the intention to deceive, is a global challenge and can only be mitigated effectively through fact-checking efforts \cite{lewandowsky2020DebunkingHandbook2020}. Generally, fact-checking is defined as the assessment of the truthfulness of a check-worthy claim \cite{vlachos2014FactCheckingTask, hassan2017AutomatedFactCheckingDetecting}. For fact-checking to be effective, fact-checking itself must be convincing and justified \cite{guo2022SurveyAutomatedFactChecking}. A well-known source of human-verified knowledge is PolitiFact\footnote{https://politifact.com}, where experts manually identify check-worthy claims from news and social media and document their verification efforts in written articles. Traditional fact-checking of these claims relies on human-driven exploration, analysis, and conclusion. Consequently, this process is rather slow and expensive, lagging behind the rapid spread of misinformation. Delayed fact-checking efforts allow false narratives to take hold, distort reality, and influence public opinion, a vulnerability that is often exploited by bad actors \cite{nyhan2020FactsMythsMisperceptions}.
\\AFC systems assist human efforts to combat misinformation by leveraging state-of-the-art techniques from areas such as Natural Language Processing (NLP), Natural Language Generation (NLG), and Information Retrieval (IR). Ideally, these systems automatically extract claims from the presented media, retrieve relevant and credible references, and provide evidence-based verdicts on the aggregated results. As opposed to style-based detection approaches that learn to distinguish claims based on writing patterns, AFC systems follow a knowledge-based approach that relies on verification knowledge to make judgements on claims \cite{zhou2021SurveyFakeNews}. Expert fact checkers can utilize AFC systems as intelligent decision support assistance to eliminate repetitive manual tasks, highlight inconsistencies, and present their findings \cite{graves2018UnderstandingPromiseLimits}. 
\\Humans often distrust fact-checking work that challenges their beliefs, perceiving them as biased or manipulated \cite{brandtzaeg2017TrustDistrustOnline}. This skepticism is likely to be aggravated with closed systems, where the lack of transparency around internal mechanisms and design decisions further erodes trust. \citet{brandtzaeg2017TrustDistrustOnline} argue that to strengthen trust, fact-checking processes must be made transparent. 
\\LLMs such as GPT-4, Claude 3.5 Sonnet, and Llama-3, have provided significant potential for a broad range of text-to-text reasoning tasks. Integrating LLMs as an inference engine into AFC systems may enhance transparency by generating veracity predictions and the accompanying natural language explanations. However, \citet{setty2024SurprisingEfficacyFineTuned} demonstrate that, for AFC-related classification tasks, fine-tuned small language models (SLM) outperform LLMs. This indicates that further research is needed to effectively utilize LLMs for AFC.
\\This paper investigates the task formulation and assessment for AFC of real-world claims with LLMs to establish baselines in various settings and to evaluate whether truthfulness ratings can be effectively modeled or if alternative approaches to claim annotations and task formulation are necessary. Based on these findings, future approaches can make more informed design choices and improve the reliability and effectiveness of AFC systems. We propose a framework for AFC with LLMs in a few-shot setup without model fine-tuning for claim analysis, claim veracity prediction, and the generation of justifications as natural language explanations. In the scope of this work, we assess the performance of our framework on 17,856 real-world claims from PolitiFact based on three labeling schemes, with or without web evidence, and across models of different sizes (i.e. 3B, 8B, 70B). Using reference-free evaluation metrics and conducting extensive experiments, we provide insight into how evidence integration, model size, and labeling complexity impact system performance. Additionally, we consider fine-tuning small state-of-the-art classification models for estimating the upper bound of predictive performance extractable from different components of the data points in the collected dataset and assess the performance of LLMs with a few-shots relative to this limit. Thus, our findings contribute to the development of more robust and transparent AFC systems using LLMs. 

\section{Related Work}

NLP provides the foundation for efficiently processing and interpreting text, making it critical for addressing misinformation detection. Advances in transformer-based architectures have significantly improved language modeling and generation. Vaswani et al. \cite{vaswani2023AttentionAllYou} introduced the transformer architecture, which leverages multi-head self-attention to capture contextual dependencies between tokens. Variants such as BERT \cite{devlin2019BERTPretrainingDeep} and GPT \cite{radford2018ImprovingLanguageUnderstanding} illustrate encoder-only and decoder-only applications, respectively. BERT is well-suited for sequence classification tasks, while GPT generates text autoregressively for sequence-to-sequence tasks. In the context of this study, we refer to these lightweight architectures as SLMs \cite{nguyen2024SurveySmallLanguage}. \\\citet{kaplan2020ScalingLawsNeural} showed that the loss of a language scales as a power law with model parameter size, pre-training dataset size, and compute budget, guiding the development of current large language models (LLMs). Most LLMs follow the GPT architecture with modifications, aligning their b ,ehavior to specific tasks through instruction fine-tuning, reinforcement learning from human feedback (RLHF) \cite{ouyang2022TrainingLanguageModelsa}, direct preference optimization (DPO) \cite{rafailov2023DirectPreferenceOptimization}, or supervised fine-tuning. Prompts serve as inputs to these models, acting as task instructions. Effective prompt engineering, including template design and task-specific examples in a few-shot scenario, enhances their utility \cite{schulhoff2024PromptReportSystematic, brown2020LanguageModelsAre}. Retrieval-Augmented Generation (RAG) \cite{lewis2021RetrievalAugmentedGenerationKnowledgeIntensive} is a technique that combines retrieval mechanisms with NLG, allowing LLMs to ground their responses in external evidence. This approach can enhance factual accuracy of LLM generated responses by enforcing consistency with the evidence, and thus may reduce hallucinations in generative tasks. Automated fact-checking frameworks typically consist of claim detection, evidence retrieval, and claim verification \cite{guo2022SurveyAutomatedFactChecking}. Claim detection identifies check-worthy claims \cite{hassan2017AutomatedFactCheckingDetecting}, often guided by factors like relevance or harm. Evidence retrieval involves collecting and selecting relevant information to justify verdicts \cite{ferreira2016EmergentNovelDataset}. Claim verification can be broken down into two main tasks: (a) verdict prediction and (b) justification production \cite{guo2022SurveyAutomatedFactChecking}. Verdict prediction describes the task of classifying the veracity of a claim. Specific labeling schemes may be required to fit particular use cases or domains. \citet{augenstein2019MultiFCRealWorldMultiDomain} investigate various datasets for knowledge-based fact-checking with different labeling schemes, ranging from a simple binary classification task formulation to more nuanced labels that do not directly map onto a veracity scale. The authors highlight the challenge of standardizing these schemes across datasets. Several approaches to justification production exist. \citet{kotonya2020ExplainableAutomatedFactChecking} survey explainable automated fact-checking, emphasizing the need for effective explanation functionality, comparing current methods against desirable properties, and highlighting existing limitations. Similarly, \citet{russo2023BenchmarkingGenerationFact} treat justification production as an abstractive summarization task using language models. However, \citet{maynez2020FaithfulnessFactualityAbstractive} note that abstractive extraction models are prone to hallucinations, underlining the necessity for more systematic approaches of generating justifications to ensure transparency and reliability. While LLMs can help produce justifications, users may over-rely on these explanations even when they are incorrect \cite{si2024LargeLanguageModels}. \citet{eldifrawi2024AutomatedJustificationProduction} explore various methods for producing justifications, such as the \textit{predict-then-justify} approach, where a classification model determines veracity followed by a separate model generating explanations, and the \textit{justify-then-predict} approach, which employs reasoning techniques like chain-of-thought \cite{pan2023FactCheckingComplexClaims} and multi-hop \cite{wang2023ExplainableClaimVerification} with LLMs.

\section{Dataset}
\label{sec:dataset}

Fact-checking organizations, that document their efforts and share them publicly, offer a great opportunity to analyze relevant misinformation and model the verification process. Moreover, by providing the initial judgment on what is check-worthy or not, fact-checking experts greatly reduce the complexity of the task at hand \cite{vlachos2014FactCheckingTask}. At PolitiFact, experts select check-worthy claims by determining whether they are verifiable as opposed to opinions and personal experiences, potentially misleading, significant enough to influence public discourse, likely to be repeated, or if a typical reader would reasonably question their truthfulness. The content at PolitiFact is localized around topics that can be found in US news. In this work, we utilize a dataset collected from PolitiFact’s online repository of fact-checking efforts. PolitiFact is a frequently used source of misinformation data, as seen in LIAR LIAR \cite{wang2017LiarLiarPants} or Mocheg \cite{yao2023EndtoEndMultimodalFactChecking}.
\\We collect 23,495 data points from English PolitiFact articles between 2007 and the 26th of January 2024. Claims not attributed to public figures (i.e. social media posts) were excluded, as these were predominantly evaluated as fake, resulting in a refined dataset of 17,856 claims. In the context of this research, we are interested in collecting the claims that have been deemed check-worthy, the entity that shared said claim, the context in which the claim has been produced, and finally the rating which has been assigned to the claim. We also match and provide the background descriptions of the entity that produced the claim. Figure \ref{politifact_datapoint} illustrates the available features.

\begin{figure}[h]
    \centering
    \fbox{%
    \begin{minipage}{0.6\textwidth}
        \textbf{Source:} New York Times Editorial Board \\
        \textbf{Background:} The editorial board is made up of 16 journalists ... \\
        \textbf{Context:} stated on June 14, 2017 in a New York Times editorial \\
        \textbf{Claim:} "A political map circulated by Sarah Palin's 2019s PAC incited Rep. Gabby Giffords's 2019 shooting" \\
        \textbf{Label:} False
    \end{minipage}
    }
    \caption{Example data point of a statement made by the New York Times Editorial Board and evaluated by PolitiFact as False.}
    \label{politifact_datapoint}
\end{figure}
\noindent PolitiFact’s rating system follows an ordinal six-class labeling scheme. Table~\ref{tab:politifact_ratings} provides the official descriptions of these six classes. 

\begin{table}[htbp]
\centering
\caption{Definitions of the original PolitiFact rating system labels.}
\label{tab:politifact_ratings}
\begin{tabular}{@{}ll@{}}
\toprule
\textbf{Label} & \textbf{Definition} \\ \midrule
\textbf{TRUE} & ... is accurate and there’s nothing significant missing. \\
\textbf{MOSTLY TRUE} & ... is accurate but needs clarification or additional information. \\
\textbf{HALF TRUE} & ... is partially accurate but leaves out important details or takes things out of context. \\
\textbf{MOSTLY FALSE} & ... contains an element of truth but ignores critical facts [...]. \\
\textbf{FALSE} & ... is not accurate. \\
\textbf{PANTS ON FIRE} & ... is not accurate (thus false) \textbf{and} makes a ridiculous claim. \\
\bottomrule
\end{tabular}
\end{table}

\noindent While PolitiFact assigns a separate \emph{PANTS ON FIRE} label to document the characteristic of ridiculousness in claims, we are only interested in the dimension of truthfulness and therefore treat this special label as a sub-case of \emph{False}. Thus, we merge the classes, discard the sixth label, and reduce the overall set of labels to five. Table \ref{all_class_dist} illustrates the resulting distribution of classes. 




\section{Methodology}

This section outlines the methodology used to design and evaluate our framework for automated fact-checking with LLMs. Following the description of the data collection from PolitiFact, we formulate the problem and the experimental setup. Specifically, we discuss model selection, labeling scheme choices, and evidence retrieval. 

\subsection{Task Formulation}

The approach in this study is motivated by the need to enhance coherence, consistency, and interpretability in automated fact-checking systems. By combining reasoning, classification, and explanation as justification within a single framework, we aim to leverage intermediate analysis to improve performance and ensure consistency between outputs. This study approaches automated fact-checking as a multi-component task with three key objectives:

\begin{enumerate}
    \item \textbf{Reasoning}: Producing a detailed, step-by-step analysis of the claim using the available information.
    \item \textbf{Verdict}: Assigning a veracity label to the claim based on a predefined set of categories.
    \item \textbf{Explanation}: Providing a clear and concise explanation in natural language to support the assigned verdict.
\end{enumerate}

\noindent The reasoning task follows the idea of chain-of-thought reasoning \cite{wei2023ChainofThoughtPromptingElicits} by constructing a step-by-step analysis of the available information as a natural language explanation \cite{kotonya2024FrameworkEvaluatingExplanations}. Thus, the verdict classification is integrated with both preceding analysis and subsequent explanation to enhance performance, building on insights from existing research. Zhang et al. \cite{zhang2021ExplainPredictThen} demonstrate that jointly generating explanations and predictions outperforms explain-then-predict models. Similarly, Atanasova et al. \cite{atanasova2020GeneratingFactChecking} find that generating fact-checking explanations alongside veracity predictions improves both the performance and the quality of the explanations. These tasks are addressed within a one-shot classification framework, utilizing instruction-based prompts to guide LLMs in generating structured outputs. 

\subsection{Prompt Design}
\label{sec:promptdesign}
We design the prompts based on the previously outlined problem formulation and established principles of prompt engineering \cite{schulhoff2024PromptReportSystematic}. Each prompt is composed of three main components: system, user, and assistant. The system message sets the model's context and provides the instructions, including the selected labeling scheme. The user message specifies the speaker, context, and claim, with evidence included when available. The assistant message contains the model's response to the input. To simulate a chat history with desired outputs, one-shot examples are included as user and assistant message pairs following the system message and preceding the actual input.

\begin{tcolorbox}[colback=gray!5!white, colframe=gray!75!black]
\textbf{SYSTEM:} You are an intelligent decision support system for automated fact-checking.\\ Your tasks are:
\begin{enumerate}
    \item Analyze the claim step-by-step.
    \item Classify the claim's veracity based on your analysis. [LABELS]
    \item Provide a concise natural language explanation for the verdict prediction.
\end{enumerate}
\end{tcolorbox}

\begin{tcolorbox}[colback=gray!5!white, colframe=gray!75!black]
\textbf{USER:} [SPEAKER][CONTEXT] the claim [CLAIM]. Evidence: [EVIDENCE]
\end{tcolorbox}

\begin{tcolorbox}[colback=gray!5!white, colframe=gray!75!black]
\textbf{ASSISTANT:} 
\end{tcolorbox}

\noindent To ensure consistency and enable automated processing, we enforce a structured output format using the \texttt{vLLM}\footnote{https://github.com/vllm-project/vllm} and \texttt{outlines}\footnote{https://github.com/dottxt-ai/outlines} libraries. 
In this context, structure refers to the property of generated output satisfying a constrained syntax \cite{willard2023EfficientGuidedGeneration}. The output is generated as a parsable JSON object with the following properties: reasoning, verdict, and explanation. Reasoning as free-text, step-by-step analysis of the claim. The verdict as the predicted veracity label, constrained to any option of the predefined set of labels. Lastly, the concise natural language explanation arguing the verdict prediction. 

\subsection{Model Selection}

To evaluate performance across different model scales, we selected a range of LLMs from the Llama 3 series. We choose Llama architecture models due to their state-of-the-art performance and open-source availability, making them well-suited for evaluating automated fact-checking systems. The models used in this study are Llama-3.2-3B, Llama-3.1-8B, Llama-3.1-70B, Llama-3.3-70B in their instruction-finetuned state. The selection covers varying parameter sizes (3B, 8B, 70B) to investigate the relationship between model scale and task performance. Our strategy is to evaluate the most recent model available at each size. The 3.2 line was the first to introduce the 3B size, while the only 8B version is found in the 3.1 line. For the 70B size, checkpoints are available in both the 3.1 and 3.3 lines. All models have a December 2023 knowledge cutoff. During pre-training, the 3.2 models processed 9 trillion tokens, whereas the 3.1 and 3.3 models processed 15 trillion tokens. The 3.3 70B Llama model achieves comparable performance to the 3.1 405B model\footnote{https://huggingface.co/meta-llama/Llama-3.3-70B-Instruct\#benchmarks}, making it one of the most performant open source models at this size. This justifies its inclusion as an additional option in model selection. All models are used in their instruction-tuned state to ensure alignment with the task. Instead of further fine-tuning, we rely on the models' available capabilities to perform one-shot reasoning, classification and explanation.

\subsection{Label Schemes}

Fact-checkers adopt varied approaches to labeling schemes, reflecting different priorities and methodologies. Some, such as FullFact\footnote{https://fullfact.org/}, rely solely on justifications without assigning explicit ratings to claims. Others, like PolitiFact and Snopes\footnote{https://snopes.com/}, implement labeling systems grounded in the idea of truthfulness. A further extension of these schemes includes labels for scenarios where evidence is incomplete or unavailable. In the AFC community, truthfulness labels are frequently mapped to a conceptual dimension that evaluates factuality based on available ground-truth evidence. Labels such as supported, refuted, cherry-picked, or NEI (not enough evidence) are commonly used \cite{hanselowski2019RichlyAnnotatedCorpus, schlichtkrull2023AVeriTeCDatasetRealworld, yao2023EndtoEndMultimodalFactChecking}, requiring significant human effort for exploration and annotation. While these approaches provide valuable insights, they also introduce complexities related to interpretation and consistency in annotations. We postpone this perspective to future work. Our focus in this study is to assess whether fact-checking can be effectively modeled across different granularities of truthfulness on the collected data. Specifically, we aim to evaluate the trade-offs between simpler and more nuanced labeling schemes in terms of their impact on classification performance and justification quality. 
\noindent To evaluate the impact of label granularity on fact-checking performance, we merge the five original PolitiFact labels (\emph{True}, \emph{Mostly True}, \emph{Half True}, \emph{Mostly False}, \emph{False}) into coarser schemes, progressively reducing complexity while preserving interpretability. In the three-class scheme, the original labels true and false are grouped into mostly true and mostly false, respectively. In the binary scheme, the label half-true is merged into mostly true. We aim to align our label aggregation with PolitiFact’s definitions, as introduced in Table \ref{tab:politifact_ratings}. Table \ref{all_class_dist} illustrates the resulting distributions.

\begin{table}[t]
    \centering
    \caption{Distribution of data for five, three, and two-class Settings}
    \label{all_class_dist}
    \tabcolsep=0.15cm
            \begin{tabular}{llccccc} 
                \toprule
                \multirow{1}{*}{\textbf{\textsc{Per-class}}} & \multicolumn{2}{c}{\textbf{5-class}} & \multicolumn{2}{c}{\textbf{3-class}} & \multicolumn{2}{c}{\textbf{2-class}} \\ 
                \textbf{Labels}  & \textbf{Count} & \textbf{Percentage}  & \textbf{Count} & \textbf{Percentage}  & \textbf{Count} & \textbf{Percentage}\\
            
                \midrule
                true  & 2531           & 14.18\% & -           & - & -  & - \\
                mostly-true        & 3347       & 18.75\% & 5878           & 32.92\%             & 9412           & 52.71\%             \\
                 half-true      & 3534         & 19.79\% & 3534           & 19.79\%    & - & -         \\ 
                 mostly-false         & 3212           & 17.99\%  & 8443           & 47.29\%  & 8443           & 47.29\%             \\ 
                 false                & 5231           & 29.30\%  & - & - & - & -\\

\bottomrule
\end{tabular}
\end{table}

\noindent We retain the PolitiFact definitions as describe in section \ref{sec:dataset} for each label across schemes.
By evaluating these schemes, we aim to understand how different levels of granularity influence the model’s ability to classify claims and provide useful explanations.

\subsection{Evidence Retrieval}

Though PolitiFact’s fact-checking articles provide human-collected evidence that informs the justification and final verdict, extracting and decontextualizing these evidences is not trivial and requires additional specialized modeling and annotation. Consequently, in this study we focus on web-based fact-checking to gather relevant information. We collect the evidence by querying a web search API\footnote{https://serper.dev/} for each claim and retrieve the top 10 search results. We do not apply any query optimization or re-ranking of results. We restrict the search to exclude a list of well-known US fact-checking sites as well as snippets that mention keywords such as "PolitiFact", "fact-check", or "debunk" to exclude fact-checking articles or direct references. This way, we aim to reduce information leaking in from pages reporting the actual verification results, rather than evidence. Due to these constrains, we were not able to retrieve evidences for 667 claims. Table \ref{tab:web_evidence_titles_snippets} lists three search results for the claim presented in Figure \ref{politifact_datapoint} where we shortened the snippet text and removed the title and source URLs.

\begin{table}[ht]
\centering
\caption{Snippets of web evidence related to Sarah Palin and the New York Times editorial case.}
\label{tab:web_evidence_titles_snippets}
\begin{tabular}{ll}
\toprule
\textbf{Date}  & \textbf{Snippet} \\ \midrule

Feb 10, 2022 & Sarah Palin testified Thursday that she felt “mortified” ... \\ \midrule

Sep 10, 2020 & Palin's political action committee circulated a map of ...  \\ \midrule

Aug 16, 2017  & The Times subsequently issued a correction stating that no such link ... \\
\bottomrule
\end{tabular}
\end{table}

\subsection{Experimental Setup}
        
To assess the performance of our automated fact-checking approach, we utilize a combination of classification and generation evaluation metrics. These metrics evaluate both the performance of verdict classification and the quality of generated outputs, ensuring a comprehensive analysis of system performance. We report accuracy and F1-Scores in different aggregations strategies to observe different aspects of the classification results. To evaluate the quality of generated outputs, we use TIGERScore, a reference-free metric that assesses text quality based on a set of criteria and assigns penalties to mistakes \cite{jiang2024TIGERScoreBuildingExplainable}. TIGERScore provides an error evaluation of the generated outputs and assigns penalty scores between $[-5, -0.5]$ for each error without relying on ground truth references. The penalty scores are added up and reported for each case. Thus, a score close to $0$ shows higher quality output. In this study, we utilize the 13B TIGERScore model with default hyperparameters to evaluate generated outputs. The evaluation prompt design follows our task prompt as described in section \ref{sec:promptdesign}.
\\Due to the stochastic nature of LLMs, evaluation is often not trivial. Thus, we run each fact-checking task three times and report the majority vote for the classification performance evaluation. Additionally, as TIGERScore is a generative metric, we also run it three times and report the average metric for the justification quality assessment.

\section{Evaluation}

The evaluation section presents a detailed analysis of our automated fact-checking approach. We assess the task performance based on model size, labeling scheme, and the impact of evidence retrieval on both classification performance and the quality of generated outputs. This evaluation is structured around our predefined hypotheses and utilizes the previously introduced range of metrics to ensure a robust assessment. Additionally, statistical analyses are conducted to determine the significance of observed performance differences. 


\subsection{Hypotheses}
\label{sec:hypotheses}

Our evaluation focuses on several fundamental questions regarding the introduced problem setting. We examine whether models can reliably distinguish between the original truthfulness labels, or if alternative approaches to claim annotation and the fact-checking task formulation are required. We also consider potential limitations on the granularity of truthfulness labels that models can effectively handle. Additionally, we assess the role of parametric knowledge in task performance, specifically whether model size yields the expected effect of better performance. Finally, we investigate the impact of evidence integration on task performance. Based on these research questions, our evaluation is structured around the following hypotheses: \\\\
\textbf{Hypothesis} $H_1$: Classification task performance decreases as label complexity increases.  \\
\textbf{Hypothesis} $H_2$: Justification quality decreases as label complexity increases.\\
\textbf{Hypothesis} $H_3$: 
Retrieving and incorporating evidence improves both classification accuracy and the quality of generated justifications.  \\
\textbf{Hypothesis} $H_4$: Larger models perform better in the classification task and produce higher quality justifications.\\
\textbf{Hypothesis} $H_5$: Smaller models benefit more significantly from evidence integration than larger models due to less parametric knowledge being available. 

\subsection{Example Output}

Previously, we introduced a claim involving the New York Times editorial and Sarah Palin in Table \ref{politifact_datapoint} and showcased examples of retrieved web evidence in Table \ref{tab:web_evidence_titles_snippets}. In Figure \ref{fig:reasoning_verdict_justification}, we now present an actual output generated by the Llama3.3-70B model under the evidence-augmented setting with the five-class labeling scheme.

\begin{figure}[ht]
\centering
\begin{tabular}{|p{0.2\textwidth}|p{0.7\textwidth}|}
\hline
\textbf{Section} & \textbf{Content} \\ \hline
\textbf{Reasoning} & The New York Times editorial board claimed that a political map circulated by Sarah Palin's PAC incited Rep. Gabby Giffords' shooting. However, there is no established link between the two events, and the Times subsequently issued a correction. The case was later brought to court, with Sarah Palin suing the New York Times for defamation. The court ultimately ruled in favor of the New York Times, stating that the editorial did not defame Palin. \\ \hline
\textbf{Verdict} & False \\ \hline
\textbf{Explanation} & The statement is not accurate. There is no established link between the political map circulated by Sarah Palin's PAC and the shooting of Rep. Gabby Giffords, and the New York Times issued a correction after publishing the claim. \\ \hline
\end{tabular}
\caption{Analysis of the New York Times editorial case involving Sarah Palin.}
\label{fig:reasoning_verdict_justification}
\end{figure}

\noindent The output in Table \ref{fig:reasoning_verdict_justification} demonstrates good justification quality. The verdict is correctly classified as False, aligning with the evidence and reasoning provided. The reasoning section effectively incorporates the retrieved evidence, presenting a detailed analysis of the claim and referencing the correction issued by the New York Times. It also mentions the court's ruling in favor of the publication, which is not directly relevant to the claim verification. The explanation is concise and supports the verdict, accurately summarizing the key points without introducing ambiguity. This example highlights the potential of retrieval-augmented generation to improve classification accuracy and justification quality.

\subsection{Results}

The results presented in Tables~\ref{tab:results-5classes},\ref{tab:results-3classes}, and \ref{tab:results-binary} illustrate the results in classification performance across different labeling schemes and model sizes, with and without evidence retrieval. For the five-class setup (Table~\ref{tab:results-5classes}), evidence retrieval consistently enhances model performance, as seen in higher $F1$ scores and TIGERScore improvements. However, the 3B model struggles to outperform the baseline significantly, indicating limited capacity in handling a complex task such as automated fact-checking.

\begin{table}[ht]
\centering
\caption{Results for 5 Classes}
\label{tab:results-5classes}
\begin{tabular}{llccccc}
\toprule
\textbf{Model} & \textbf{Evidence} & \textbf{$F1_{macro}$} & \textbf{$F1_{weighted}$} & \textbf{$F1_{micro}$} & Acc. & TIGER$\uparrow$ \\
\midrule
\multirow{1}{*}{Baseline} 
        & - & 0.2 & 0.213 & 0.213 & 0.213 &  -\\
\midrule
\multirow{2}{*}{3.2-3B-Instruct} 
        & No & 0.216 & 0.242 & 0.273 & 0.273 & -3.995\\
        & Yes & 0.259 & 0.285 & 0.321 & 0.321 & -3.116\\
\midrule
\multirow{2}{*}{3.1-8B-Instruct} 
        & No & 0.244 & 0.268 & 0.293 & 0.293 & -3.416\\
        & Yes & 0.278 & 0.301 & 0.339 & 0.339 & -2.578\\
\midrule
\multirow{2}{*}{3.1-70B-Instruct} 
        & No & 0.314 & 0.338 & 0.356 & 0.356 & -2.554 \\
        & Yes & 0.335 & 0.359 & 0.389 & 0.389 & -2.466\\
\midrule
\multirow{2}{*}{3.3-70B-Instruct} 
        & No & 0.314 & 0.337 & 0.357 & 0.357 & -2.361\\
        & Yes & 0.351 & 0.375 & \textbf{0.405} & 0.405 & \textbf{-1.686} \\
\bottomrule
\end{tabular}
\end{table}

\noindent In the three-class classification scheme (Table~\ref{tab:results-3classes}), evidence retrieval again provides a notable performance boost across all models, with improvements becoming more pronounced in larger models. This indicates that as label complexity decreases, models are better able to leverage evidence to enhance classification accuracy and justifications. The 3.3-70B-Instruct model achieves the highest scores, emphasizing the advantage of size when combined with external knowledge. Table~\ref{tab:results-3classes} presents the classification metrics for the three-class scheme. Similar to the five-class results, evidence retrieval enhances performance across all models.

\begin{table}[ht]
\centering
\caption{Results for 3 Classes}
\label{tab:results-3classes}
\begin{tabular}{llccccc}
\toprule
\textbf{Model} & \textbf{Evidence} & \textbf{$F1_{macro}$} & \textbf{$F1_{weighted}$} & \textbf{$F1_{micro}$} & Acc. & TIGER$\uparrow$ \\
\midrule
\multirow{1}{*}{Baseline} 
        &  & 0.333 & 0.371 & 0.371 & 0.371 & - \\
\midrule
\multirow{2}{*}{3.2-3B-Instruct} 
        & No & 0.322 & 0.396 & 0.464 & 0.464 & -4.069\\
        & Yes & 0.390 & 0.453 & 0.498 & 0.498 & -3.205\\
\midrule
\multirow{2}{*}{3.1-8B-Instruct} 
        & No & 0.389 & 0.443 & 0.472 & 0.472 & -3.391\\
        & Yes & 0.448 & 0.506 & 0.525 & 0.525 & -2.751\\
\midrule
\multirow{2}{*}{3.1-70B-Instruct} 
        & No & 0.499 & 0.551 & 0.542 & 0.542 & -2.610\\
        & Yes & 0.521 & 0.571 & 0.556 & 0.556 & -2.524\\
\midrule
\multirow{2}{*}{3.3-70B-Instruct} 
        & No & 0.524 & 0.570 & 0.556 & 0.556 & -2.383\\
        & Yes & 0.550 & 0.601 & \textbf{0.589} & 0.589 & \textbf{-1.884}\\
\bottomrule
\end{tabular}
\end{table}

\noindent For binary classification results in Table~\ref{tab:results-binary}, the reduced complexity of the task yields the highest overall performance across all models. Evidence retrieval continues to provide a measurable benefit, particularly in the largest models, where the highest $F1$ scores and TIGERScore improvements are observed.

\begin{table}[ht]
\centering
\caption{Results for Binary Classification}
\label{tab:results-binary}
\begin{tabular}{llccccc}
\toprule
\textbf{Model} & \textbf{Evidence} & \textbf{$F1_{macro}$} & \textbf{$F1_{weighted}$} & \textbf{$F1_{micro}$} & Acc. & TIGER$\uparrow$ \\
\midrule
\multirow{1}{*}{Baseline} 
        & - & 0.500 & 0.501 & 0.501 & 0.501 & - \\
\midrule
\multirow{2}{*}{3.2-3B-Instruct} 
        & No & 0.624	 & 0.624 & 0.624 & 0.504 & -3.870\\
        & Yes & 0.647 & 0.647 & 0.647 & 0.557 & -3.150 \\
\midrule
\multirow{2}{*}{3.1-8B-Instruct} 
        & No & 0.649 & 0.649 & 0.649 & 0.543 & -3.367  \\
        & Yes & 0.668 & 0.668 & 0.668 & 0.589 & -2.741\\
\midrule
\multirow{2}{*}{3.1-70B-Instruct} 
        & No & 0.689 & 0.689 & 0.689 & 0.684 & -2.560\\
        & Yes & 0.708 & 0.708 & 0.708 & 0.691 & -2.433\\
\midrule
\multirow{2}{*}{3.3-70B-Instruct} 
        & No & 0.722 & 0.722 & 0.722 & 0.707 & -2.303\\
        & Yes & 0.747 & 0.747 & \textbf{0.747} & 0.734 & \textbf{-1.739}\\
\bottomrule
\end{tabular}
\end{table}

\noindent In the following, we investigate the indications we have described earlier with statistical analyses to draw conclusions about the hypotheses we specified in section \ref{sec:hypotheses}.
\\We conducted a Friedman test on $F1_{micro}$ across the three classification schemes, with and without evidence. The result indicates that at least one of the schemes differs significantly ($p < 0.05$) in terms of classification performance. These findings support hypothesis $H_1$ that as labeling becomes more complex, classification performance tends to decrease, potentially due to more nuanced distinctions between labels that increase the quantity of prediction errors. 
\\For TIGERScore evaluation, the Friedman test was significant ($p < 0.05$) for the setting with evidences, but a subsequent Conover's test revealed no significant pairwise differences. Additionally, the Friedman test reveals no significance for the setting without evidence. This result suggests that there is no measurable difference in justification quality across the three schemes, with or without evidence. These findings reject hypothesis $H_2$ that more complex label sets negatively affect overall justification quality. This may be in part due to claim analysis and explanation being difficult enough, regardless of whether the label schemes are more or less complex. 
\\To determine the statistical significance of performance when including evidence, we conducted paired t-tests comparing models with and without evidence across all classification schemes for the $F1_{micro}$ and TIGERScore. The results indicate a statically significant difference (p < 0.01) for both metrics when evidence retrieval is included during the fact-checking task. This supports hypothesis $H_3$. Thus, external evidence helps the model to disambiguate classes and produce more useful justifications for the fact-checking task.
\\To evaluate whether larger models outperform their smaller counterparts, we performed a Friedman test on both $F1_{micro}$ and TIGERScore with and without evidence across four different model sizes. The results indicate a significant difference ($p < 0.05$), confirming that model size has a measurable impact on performance. These findings support hypothesis $H_4$.
\\Finally, to investigate whether smaller models benefit more from evidence integration than larger models, we examined the performance gains by subtracting the no evidence scores from the with evidence scores for both $F1_{micro}$ and TIGERScore across all model sizes. For $F1_{micro}$ gains, the Friedman test showed no significant difference ($p = 0.167$), whereas the TIGERScore gains were statistically significant ($p < 0.05$). Thus, we partially reject hypothesis $H_5$. This implies that larger models benefit even more from external evidence, presumably due to their ability to reason effectively across long context sizes, whereas smaller models exhibit relatively limited improvements. We expect that integrating more credible and complete information sources, could enhance overall performance for both smaller and larger models even further. 

\noindent Since we consider fine-tuning-based approaches impractical due to the dynamic and rapidly evolving nature of misinformation, which limits the effectiveness of models fine-tuned on static datasets, our primary focus in this study has been on the few-shot inference scenario. Earlier encoder-based architectures, such as BERT, were restricted to processing contexts of up to 512 tokens. However, recent advancements like ModernBERT \cite{warner2024SmarterBetterFaster} now enable efficient handling of sequence classification tasks with extended contexts of up to 8192 tokens. This innovation facilitates the integration of evidence directly into the classification process, enhancing its utility for predictive analysis.

\noindent To estimate the upper bound of predictive performance from the available data and to assess our models' performance relative to this limit, we employ fine-tuning with the ModernBERT-large architecture, which holds 395M trainable parameters. Our approach involves training a model for each of the following sequence classification configurations. First, we only provide the claim. Next, we incorporate the claim's context (e.g., interview or speech) as additional input. We then include the source entity responsible for the claim. Finally, we augment the input with relevant evidence.

\begin{table}[ht]
\centering
\caption{Results for fine-tuning a ModernBERT-large classifier upon different combinations of the model input.}
\label{tab:results-romcir2}
\begin{tabular}{llcccc}
\toprule
\textbf{Labels} & \textbf{Setting} & \textbf{$F1_{macro}$} & \textbf{$F1_{weighted}$} & \textbf{$F1_{micro}$} & \textbf{Accuracy} \\
\midrule
\multirow{4}{*}{5 Classes} 
        & Claim      & 0.296 & 0.293 & 0.239 & 0.296 \\
        & + Context     & 0.338 & 0.339 & 0.276 & 0.338 \\
        & + Speaker    & 0.350 & 0.354 & 0.287 & 0.350 \\
        & + Evidence   & 0.392 & 0.391 & \textbf{0.317} & 0.392 \\
\midrule
\multirow{4}{*}{3 Classes} 
        & Claim      & 0.499 & 0.462 & 0.363 & 0.499 \\
        & + Context     & 0.499 & 0.474 & 0.375 & 0.499 \\
        & + Speaker    & 0.530 & 0.527 & 0.445 & 0.530 \\
        & + Evidence   & 0.559 & 0.579 & \textbf{0.491} & 0.559 \\
\midrule
\multirow{4}{*}{2 Classes} 
        & Claim      & 0.666 & 0.664 & 0.637 & 0.666 \\
        & + Context     & 0.636 & 0.636 & 0.612 & 0.636 \\
        & + Speaker    & 0.652 & 0.653 & 0.624 & 0.652 \\
        & + Evidence   & 0.718 & 0.719 & \textbf{0.696} & 0.718 \\
\bottomrule
\end{tabular}
\end{table}

The results highlight that adding evidence consistently provides the most significant performance gains across all label schemes. In the 5-class setting, the model achieves the lowest performance with claims alone, improving slightly with context and speaker information. Evidence integration yields the highest gains. In the 3-class setting, context adds no value, while speaker information increases performance. Evidence again provides the largest improvement. For 2-class classification, claims alone achieve high baseline performance. Context slightly reduces performance, likely due to issues stemming from the label aggregation process, where claims categorized within the binary label disagree on the contexts they have been used in, introducing noise into the process. Speaker information adds moderate gains. Evidence leads to the best results. Overall, evidence is the most impactful input, with speaker information providing moderate benefits. 
\\The classification results align closely with the previously presented LLM results, showing that evidence consistently provides the most significant performance improvements across all label schemes. For both classifiers and LLMs, the inclusion of evidence enables better disambiguation and enhances predictive performance, particularly in more complex multi-class tasks. While smaller LLMs provide comparable performance across tasks in one-shot inference, larger LLMs consistently surpass the fine-tuned SLMs without requiring fine-tuning and provide the additional advantage of generating coherent reasoning and detailed justifications across more extensive contexts. This highlights the general utility of LLMs for AFC.

\section{Discussion and Conclusion}
This study investigated AFC of real-world claims using LLMs in a few-shot inference scenario. By evaluating task performance across three labeling schemes and multiple LLM sizes of the same architecture, we demonstrated the importance of evidence integration, model size, and labeling complexity in determining system effectiveness. Evidence retrieval consistently improved classification accuracy and justification quality, with larger models showing the most significant gains. Smaller models were not able to perform or benefit as well from evidence integration, emphasizing the need for further optimizing the modeling process in computationally constrained environments. While coarse-grained labels naturally yield higher performance, further research must determine how to integrate a more nuanced assessment of claims across different perspectives for a robust AFC approach. 

\noindent To extend AFC toward intelligent decision assistance for expert fact-checkers, future work should focus on structuring justifications to better align with human verification strategies. This includes presenting concise, faithful explanations in a format that describes the key reasoning steps and highlights integrated evidence. Nonetheless, initial observations indicate that LLM-based systems can suffer from hallucinations, emphasizing the need for more extensive evaluation. Additionally, exploring user studies to assess how fact-checkers interpret and trust generated explanations will provide insights into designing more useful and actionable outputs. Ensuring explanations are both transparent and structured in ways that facilitate quick validation by experts could significantly enhance system utility in professional settings. Such comparisons will help identify gaps in AFC systems and refine their role in supporting, rather than replacing, expert-driven fact-checking efforts. A valuable direction for future research involves benchmarking AFC systems against the fact-checking efforts already present in the fact-checking articles. By leveraging insights from human workflows and integrating interactive components for validation and feedback, AFC systems can evolve into effective, collaborative tools for combating misinformation.

\begin{acknowledgments}
This study is partially funded by the German Federal Ministry of Education and Research (BMBF, reference: 03RU2U151C) in the scope of the research project \href{https://news-polygraph.com/}{news-polygraph} and by JST AIP Acceleration Research (JPMJCR24U3) and JST CREST Grants (JPMJCR20D3).
\end{acknowledgments}

\bibliography{filtered_references}


\end{document}